\documentclass[journal]{IEEEtran}
\usepackage{amsfonts}
\usepackage{mathrsfs}
\usepackage{cite}
\usepackage{graphicx}
\usepackage{amsmath}
\usepackage{algorithmic}
\usepackage{algorithm}
\usepackage{array}
\usepackage{multirow}
\usepackage{booktabs}

\begin{document}

\title{Zero-Shot Fine-Grained Classification by Deep Feature Learning with Semantics}

\author{Aoxue~Li,~Zhiwu~Lu,~Liwei~Wang,~Tao Xiang,~Xinqi~Li,~and~Ji-Rong~Wen%
\thanks{A. Li and L. Wang are with the Key Laboratory of Machine Perception (MOE), School of Electronics Engineering and Computer Science, Peking University, Beijing 100871, China (email: lax@pku.edu.cn,~wanglw@cis.pku.edu.cn).}
\thanks{Z. Lu, X. Li and J.-R. Wen are with the Beijing Key Laboratory of Big Data Management and Analysis Methods, School of Information, Renmin University of China, Beijing 100872, China (email:
luzhiwu@ruc.edu.cn,~lixinqi96@163.com,~jrwen@ruc.edu.cn).}
\thanks{T. Xiang is with the School of Electronic Engineering and Computer Science, Queen Mary University of London, Mile End Road, London E1 4NS, United Kingdom (email: t.xiang@qmul.ac.uk).}}

\maketitle

\begin{abstract}

Fine-grained image classification, which aims to distinguish images with subtle distinctions, is a challenging task due to two main issues: lack of sufficient training data for every class and difficulty in learning discriminative features for representation. In this paper, to address the two issues, we propose a two-phase framework for recognizing images from unseen fine-grained classes, i.e. zero-shot fine-grained classification. In the first feature learning phase, we finetune deep convolutional neural networks using hierarchical semantic structure among fine-grained classes to extract discriminative deep visual features. Meanwhile, a domain adaptation structure is induced into deep convolutional neural networks to avoid domain shift from training data to test data. In the second label inference phase, a semantic directed graph is constructed over attributes of fine-grained classes. Based on this graph, we develop a label propagation algorithm to infer the labels of images in the unseen classes. Experimental results on two benchmark datasets demonstrate that our model outperforms the state-of-the-art zero-shot learning models. In addition, the features obtained by our feature learning model also yield significant gains when they are used by other zero-shot learning models, which shows the flexility of our model in zero-shot fine-grained classification.

\end{abstract}

\IEEEpeerreviewmaketitle

\section{Introduction}

\IEEEPARstart{F}{ine-grained} image classification, which aims to
recognize subordinate level categories, has emerged as a popular
research area in the computer vision community
\cite{Berg13,Liu12,Farrell11,Branson14}. Different from general
image recognition such as scene or object recognition, fine-grained
image classification needs to explicitly distinguish images with
subtle difference, which actually involves the classification of many subclasses of objects belonging to the same class such as birds
\cite{CUB-200-2011,CUB-200,Birdsnap}, dogs \cite{DOG} and plants
\cite{Flower,Flower2}.

\begin{figure}[t]
\vspace{0.03in}
\begin{center}
\includegraphics[width=0.97\columnwidth]{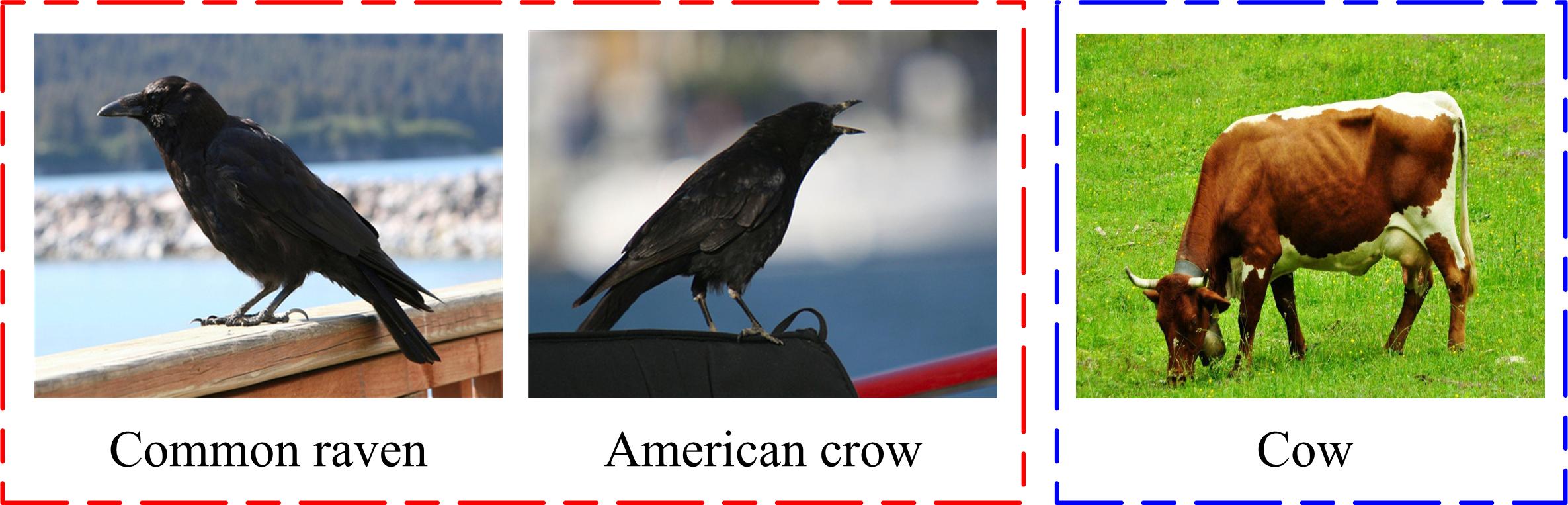}
\end{center}
\vspace{-0.05in} \caption{Fine-grained classification vs. general
image classification. Fine-grained classification (red box)
processes visually similar objects, e.g., to recognize American crow
and Common raven. General image classification usually distinguishes
an object such as birds (red box) from other objects that are
visually very different (e.g., a cow). } \label{fig1}
\vspace{-0.0in}
\end{figure}

In general, fine-grained image classification is a challenging task
due to two main issues:
\begin{itemize}
\item
Since recognizing images in the fine-grained classes is a
fairly difficult and expertise task, the annotations of images in
fine-grained classes are expensive and collecting large-scale
labelled data just as general image recognition (e.g. ImageNet
\cite{Russakovsky}) is thus impractical. Therefore, how to recognize images from fine-grained classes in the lack of sufficient training data for every class becomes a thought-provoking task in computer vision.
\item
As compared with general image recognition, fine-grained classification is a more challenging task, which needs to discriminate between objects that are visually similar to each other. As shown in Fig.\ref{fig1}, people can easily recognize that objects in the red box are birds and the object in the blue box is a cow, but they fail to distinguish the two kinds of birds in the red box. This example demonstrates that we have to learn more discriminative representation for fine-grained classification than that for general image classification.
\end{itemize}

\begin{figure*}[t]
\vspace{0.03in}
\begin{center}
\includegraphics[width=0.97\textwidth]{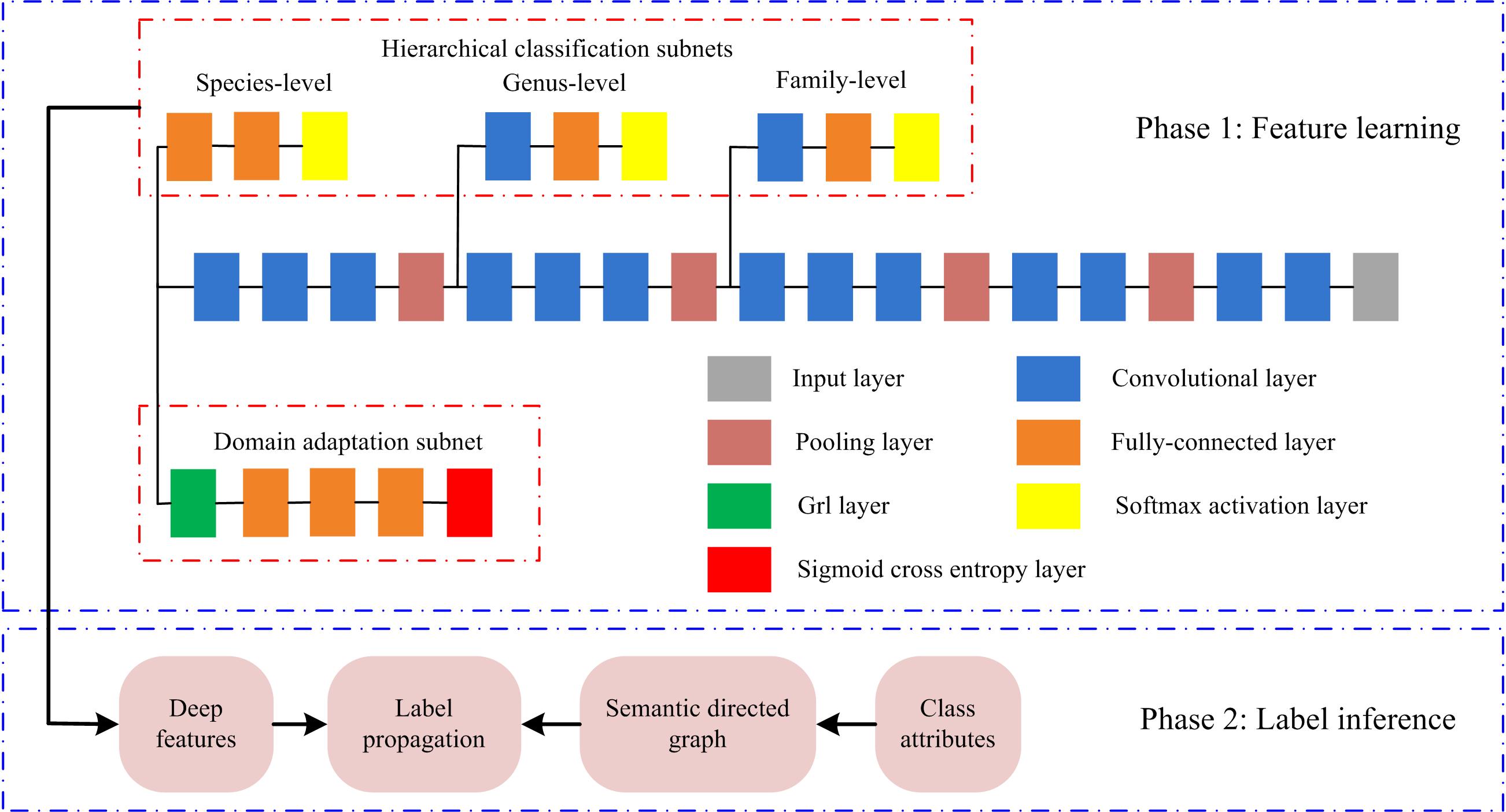}
\end{center}
\vspace{-0.05in} \caption{Overview of the proposed framework for
zero-shot fine-grained image classification. The proposed framework
contains two phases: feature learning and label inference. In the
first feature learning phase, hierarchical classification
subnetworks and a domain adaptation structure are both integrated
into VGG-16Net \cite{Simonyan15}. In the second label inference
phase, deep features from the first phase and a semantic directed
graph constructed with class attributes are involved into a label propagation process to infer the labels of images in the unseen classes.}
\label{flowchart} \vspace{-0.0in}
\end{figure*}

Considering the lack of training data for every class in fine-grained
classification, we can adopt zero-shot learning to recognize images
from unseen classes without labelled training data. However,
conventional zero-shot learning algorithms mainly explore the
semantic relationship among classes (using textual information) and
attempt to learn a match between images and their textual
descriptions \cite{Paredes15,Zhang15,Fu15}. In other words, rare
works on zero-shot learning focus on feature learning. This is
really bad for fine-grained classification, since it requires more
discriminative features than general image recognition. Hence, we
must pay our main attention to feature leaning for zero-shot
fine-grained image classification.

In this paper, we propose a two-phase framework to recognize images
from unseen fine-grained classes, i.e. zero-shot fine-grained
classification (ZSFC). The first phase of our model is to learn
discriminative features. Most fine-grained classification models
extract features from deep convolutional neural networks that are
finetuned by images with extra annotations (eg. bounding box of
objects and part locations). However, these extra annotations of
images are expensive to access. Different from these models, our
model only exploits implied hierarchical semantic structure among
fine-grained classes for finetuning deep networks. The hierarchical
semantic structure among classes is obtained based on taxonomy, which can be easily collected from Wikipedia. In our model, we generally assume that experts recognize objects in fine-grained classes based on the discriminative visual features of images and the hierarchical
semantic structure among fine-grained classes is their prior
knowledge. Under this assumption, we finetune deep convolutional
neural networks using hierarchical semantic structure among
fine-grained classes to extract discriminative deep visual features.
Meanwhile, a domain adaptation subnetwork is introduced into the
proposed network to avoid domain shift caused by zero-shot setting.

In the second label inference phase, a semantic directed graph is
firstly constructed over attributes of fine-grained classes. Based
on the semantic directed graph and also the discriminative features
obtained by our feature learning model, we develop a label propagation
algorithm to infer the labels of images in the unseen classes. The
flowchart of the proposed framework is illustrated in Fig.
\ref{flowchart}. Note that the proposed framework can be extended to weakly supervised setting by replacing class attributes with semantic
vectors extracted by word vector extractors (e.g. Word2Vec
\cite{word2vec}).

To evaluate the effectiveness of the proposed model, we conduct
experiments on two benchmark fine-grained image datasets (i.e.
Caltech UCSD Birds-200-2011 \cite{CUB-200-2011} and Oxford
Flower-102\cite{Flower}). Experimental results demonstrate that the
proposed model outperforms the state-of-the-art zero-shot
learning models in the task of zero-shot fine-grained
classification. Moreover, we further test the features extracted by our feature learning model by applying them to other zero-shot learning models and the obtained significant gains verify the effectiveness of our feature learning model.

The main contributions of this work are given as follows:
\begin{itemize}
\item
We have proposed a two-phase learning framework for
zero-shot fine-grained classification. Unlike most of previous works that focus on zero-shot learning, we pay more attention to feature learning instead.
\item
We have developed a deep feature learning method for
fine-grained classification, which can learn discriminative features with hierarchical semantic structure among classes and a domain adaptation structure. More notably, our feature learning method needs no extra annotations of images (e.g. part locations and bounding boxes of objects), which means that it can be readily used for different zero-shot fine-grained classification tasks.
\item
We have developed a zero-shot learning method for label
inference from seen classes to unseen classes, which can help to
address the issue of lack of labelled training data in fine-grained image classification.
\end{itemize}

The remainder of this paper is organized as follows. Section
\uppercase\expandafter{\romannumeral2} provides related works of
fine-grained classification and zero-shot learning. Section
\uppercase\expandafter{\romannumeral3} gives the details of the
proposed model for zero-shot fine-grained classification. Experimental results are presented in Section \uppercase\expandafter{\romannumeral4}. Finally, the conclusions are drawn in Section \uppercase\expandafter{\romannumeral5}.

\section{Related Works}

\subsection{Fine-Grained Image Classification}

There are two strategies widely used in existing fine-grained image
classification algorithms. The idea of the first strategy is
distinguishing images according to the unique properties of object
parts, which encourages the use of part-based algorithms that rely
on localizing object parts and assigning them detailed attributes.
Zhang \emph{et al.} propose a part-based Region based-Convolutional
Neural Network (R-CNN) where R-CNN is used to detect object parts
and geometric relations among object parts are used for label
inference \cite{Zhang14}. Since R-CNN extracts too many proposals
for each image, this algorithm is time-consuming. To solve this
problem, Huang \emph{et al.} propose a Part-Stacked Convolutional
Neural Network (PS-CNN) \cite{Huang16}, where a fully-convolutional
network is used to detect object parts and a part-crop layer is
induced into AlexNet \cite{Krizhevsky} to combine part/object
features for classification. To solve the limited scale of
well-annotated data, Xu \emph{et al.} propose an agumented
part-based R-CNN to utilize the weak labeled data from web
\cite{Xu15}. Different from these models that mainly use large parts
of images (i.e. proposals) for fine-grained classification, Zhang
\emph{et al.} detect semantic part and classify images based on
features of their semantic parts \cite{Zhang162}. However, the
aforementioned part-based algorithms need very strong annotations
(i.e. locations of parts), which are very expensive to acquire.

The second strategy is to exploit more discriminative visual
representations, which is inspired by recent success of CNNs in
image recognition \cite{CSzegedy}. Lin \emph{et al.} propose a
bilinear CNN \cite{Lin15}, which combines the outputs of two
different feature extractors by using a outer product, to model
local pairwise feature interactions in a translationally invariant
manner. This structure can create robust representations and achieve
significant improvement compared with the state-of-the-arts. Zhang
\emph{et al.} propose a deep filter selection strategy to choose
suitable deep filters for each kinds of parts \cite{Zhang163}. With
the suitable deep filters, they can detect more accurate parts and
extract more discriminative features for fine-grained
classification.

Note that the above models need extra annotations of images (eg. bounding boxes of objects and locations of parts). Moreover, their training data include all fine-grained classes. When we only have training images from a subset of fine-grained classes, the domain shift problem will occur \cite{Kodirov15}. Besides, without extra object or part annotations, these models will fail. In contrast, our model needs not extra object or part annotations at both training and testing stages. Furthermore, the domain adaptation strategy is induced into our model to avoid domain shift. In this way, we can learn more discriminative features for zero-shot fine-grained classification.

\begin{figure}[t]
\vspace{0.03in}
\begin{center}
\includegraphics[width=0.92\columnwidth]{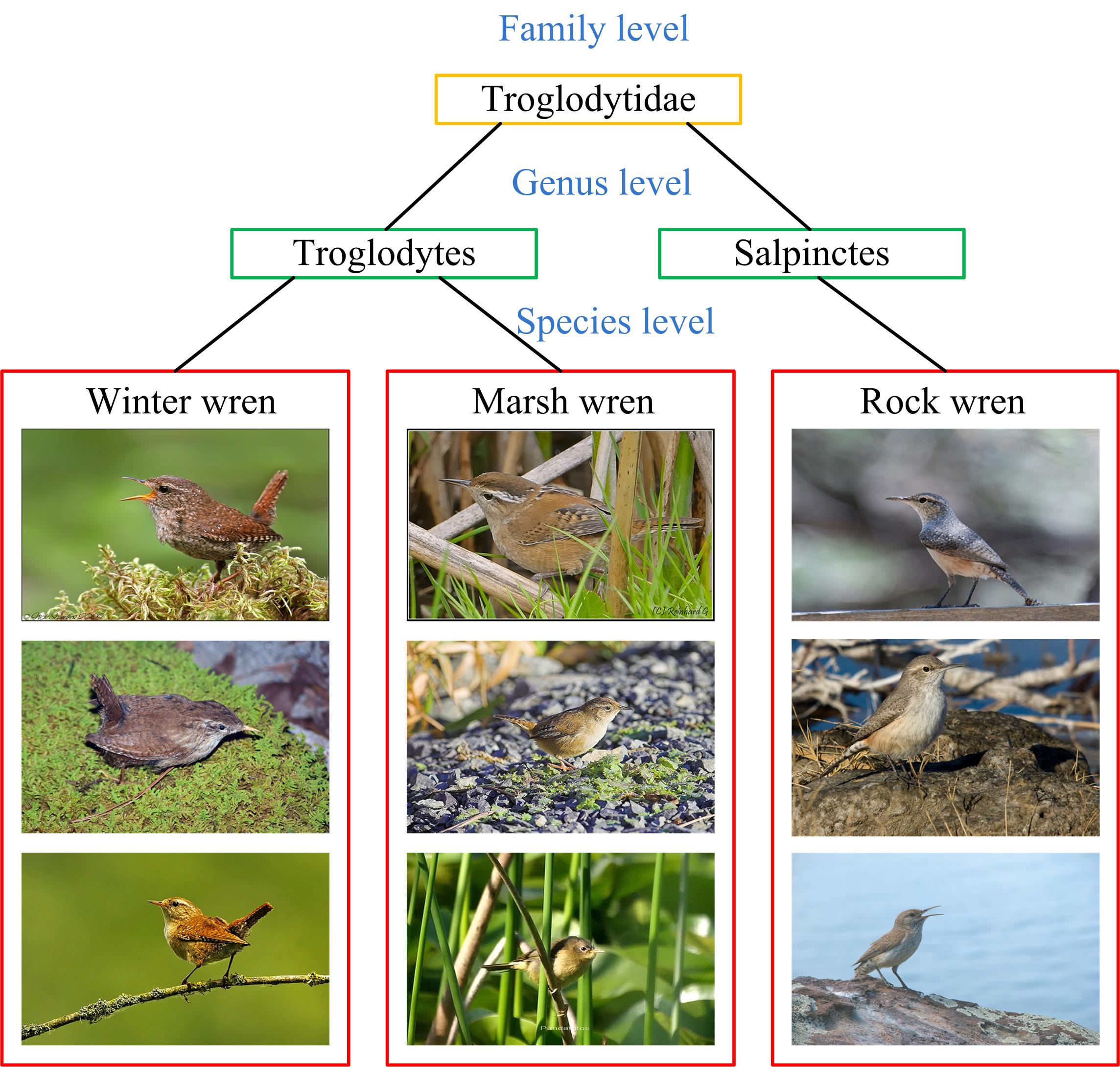}
\end{center}
\vspace{-0.1in}
\caption{Hierarchical semantic structure of fine-grained classes.}
\label{fig2}
\end{figure}

\subsection{Zero-Shot Learning}

Zero-shot learning, which aims to learn to classify in the absence
of labeled data, is a challenging problem
\cite{Lampert14,Kankuekul12,Rohrbach11,Yu10,Palatucci09,Lampert09}.
Recently, many approaches have been developed for zero-shot
learning. Zhang \emph{et al.} viewed testing instances as arising
from seen instances and attempted to express test instances as a
mixture of seen class proportions \cite{Zhang15}. To solve this
problem, they propose a semantic similarity embedding (SSE) approach
for zero-shot learning. Besides, they also formulate zero-shot
learning as a binary classification problem and develop a joint
discriminative learning framework based on dictionary learning to
solve it \cite{Zhang16}. Paredes \emph{et al.} propose a general
zero-shot learning framework to model the relationships between
features, attributes, and classes as a two linear layers network
\cite{Paredes15}. Bucher \emph{et al.} address the task of
zero-shot learning by formulating this problem as a metric learning
problem, where a metric among class attributes and image visual
features is learned for inferring labels of test images
\cite{Bucher16}. A multi-cue framework facilitates a joint embedding
of multiple language parts and visual information into a joint space
to recognize images from unseen classes \cite{Akata16}. Considering
the manifold structure of semantic categories, Fu \emph{et al.}
provide a novel zero-shot learning approach by formulating a
semantic manifold distance among testing images and unseen classes
\cite{Fu15}. To avoid domain shift between the sets of seen classes
and unseen classes, Kodirov \emph{et al.} propose a zero-shot
learning method based on unsupervised domain adaptation
\cite{Kodirov15}. On the observation that textual descriptions are
noisy, Qiao \emph{et al.} propose an $L_{2,1}$-norm based objective
function to suppress the noisy signal in the text and learn a
function to match the text document and visual features of images
\cite{Qiao16}. However, the aforementioned works mainly focus on
learning a match between images and their textual descriptions
and few of them pay attention to discriminative feature learning,
which is very crucial for fine-grained classification.

\section{The Proposed Model}

In this section, we propose a two-phase framework for zero-shot
fine-grained classification. A deep convolutional neural network
integrating hierarchical semantic structure of classes and domain
adaptation strategy is first developed for feature learning and a
label propagation method based on semantic directed graph is further
proposed for label inference.

\subsection{Feature Learning}
\label{sect:fea_learn}

\begin{figure*}[t]
\vspace{0.03in}
\begin{center}
\includegraphics[width=0.97\textwidth]{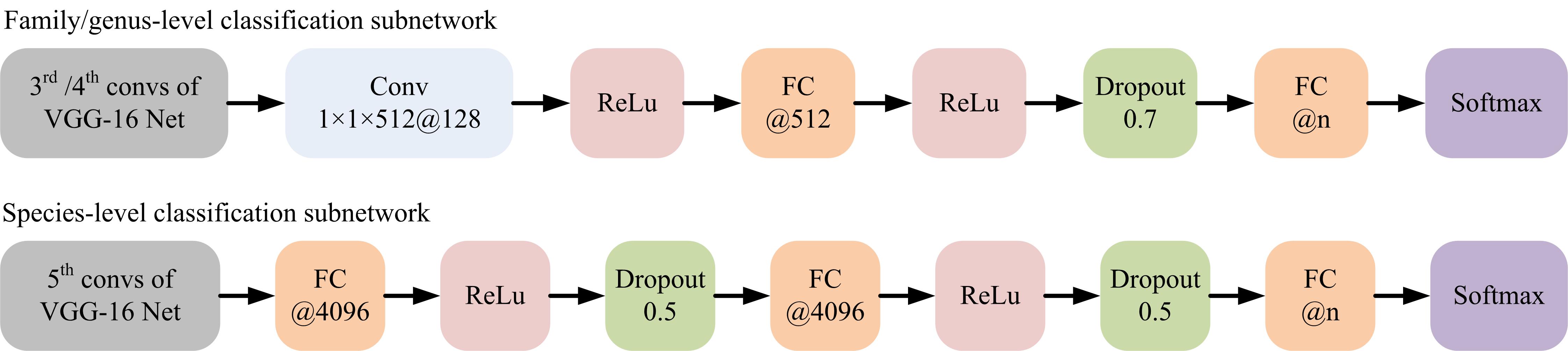}
\end{center}
\vspace{-0.05in}
\caption{Detailed architecture of hierarchical classification
subnetworks. In this figure, `Conv' and `FC' denote the
convolutional layer and fully-connected layer respectively. The
numbers under the `Conv', `FC' and `Dropout' denote the kernel
information of the convolutional layer, number of output of
fully-connected layer and the ratio of dropout, respectively. $n$ is
the total number of classes at the corresponding level.} \label{hcs}
\end{figure*}

\begin{figure*}[t]
\vspace{0.05in}
\begin{center}
\includegraphics[width=0.97\textwidth]{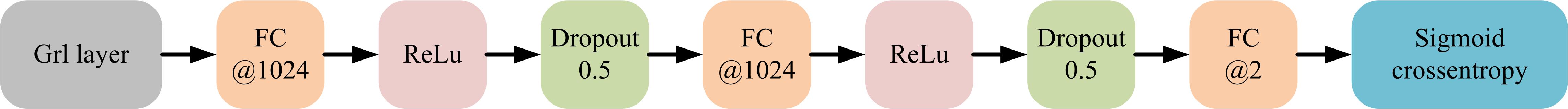}
\end{center}
\vspace{-0.05in}
\caption{Detailed architecture of domain classifier. In this figure,
`Grl' and `FC' denote gradient reversal layer and the
fully-connected layer respectively. The numbers under the `FC' and
`Dropout' denote the number of output of fully-connected layer and
the ratio of dropout, respectively.} \label{dc}
\end{figure*}

Our main idea is motivated by implied hierarchical semantic structure among fine-grained classes. For example, winter wren (species-level name), a very small North American bird, can be called `Troglodytes' at genus level and also can be called `Troglodytidae' at family level
(See Fig. \ref{fig2}). We assume that experts recognize objects in
fine-grained classes by using the discriminative visual features and
the hierarchical semantic structure among fine-grained classes is
their prior knowledge. As shown in Fig. \ref{flowchart}, lower-level features are used (with fewer network layers) for classifying images at coarser level. In other words, to recognize images in a fine-grained level, we must exploit higher-level and fine-grained features.

To induce the hierarchical semantic structure into feature learning,
we integrate hierarchical classification subnetworks into VGG-16Net
\cite{Simonyan15}. The detailed architectures of hierarchical
classification subnetworks are presented in Fig. \ref{hcs}. In our
model, each classification subnetwork is designed to classify images
into the corresponding level semantic classes (i.e. family level,
genus level, or species level). Concretely, we locate the
classification subnetworks for family-level, genus-level, and
species-level labels afterwards the third, forth, and fifth groups
of convolutional layers, respectively (also see Fig.
\ref{flowchart}). For family-level and genus-level classification
subnetworks, their detailed network structure includes a
convolutional layer, two fully-connected layers, and a softmax
activation layer (see Fig. \ref{hcs}). For the sake of quick
converegence, we take the classification structure of VGG-16Net as
the species-level classification subnetwork, which can be initialized by ImageNet pretrained parameters \cite{Russakovsky}. By merging the VGG-16Net and hierarchical classification subnetworks into one network, we define the loss function for image $x$ as:
\begin{equation}
\begin{split}
\mathcal {L}_h(\theta_F,\theta_f,\theta_g,\theta_s)= &\mu_f\mathcal{L}_f(y_f,G_f(G(x;\theta_F);\theta_f))+\\
&\mu_g\mathcal{L}_g(y_g,G_g(G(x;\theta_F);\theta_g))+\\
&\mathcal{L}_s(y_s,G_s(G(x;\theta_F);\theta_s))
\end{split}
\end{equation}
where $\mathcal {L}_f$, $\mathcal {L}_g$, and $\mathcal {L}_s$
denote the loss of family, genus, and species-level classification
subnetworks, respectively. $y_f$, $y_g$, and $y_s$ denote the true label of the image at family, genus, and species level, respectively. $\theta_F$ denotes the parameters of the feature extractor
(the first fifth groups of convolutional layers) in VGG-16Net.
$\theta_f$, $\theta_g$, and $\theta_s$ denote the parameters of
family, genus, and species-level classification subnetworks, respectively. $\mu_f$ and $\mu_g$ respectively denote the weights of loss of family and genus-level classification subnetworks. $G$ and $G_f$ (or $G_g$, $G_s$) respectively denote the feature extractor of VGG-16Net, family (or genus, species)-level hierarchical classification subnetworks.

\begin{figure*}[t]
\vspace{0.03in}
\begin{center}
\includegraphics[width=0.97\textwidth]{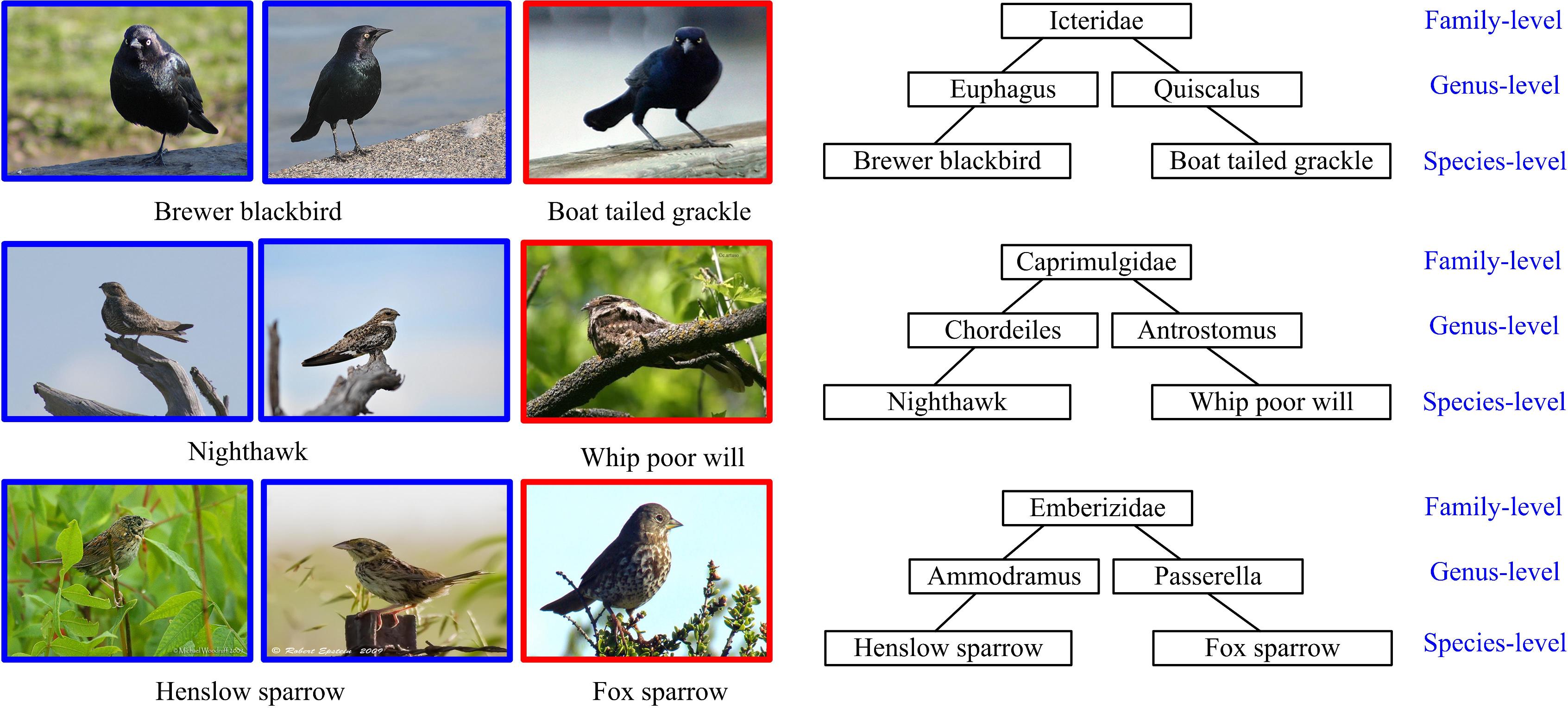}
\end{center}
\vspace{-0.05in}
\caption{Samples of misclassification only with species-level features. In this figure, images in the blue boxes are misclassified images only with species-level features and the class names under images are their true labels, while the predicted labels of these images (in blue boxes) are given by a sample (in red boxes) in the corresponding rows. These misclassified images are correctly classified when both species/genus-level features are used.}
\label{mis1}
\end{figure*}

\begin{figure*}[t]
\vspace{0.08in}
\begin{center}
\includegraphics[width=0.97\textwidth]{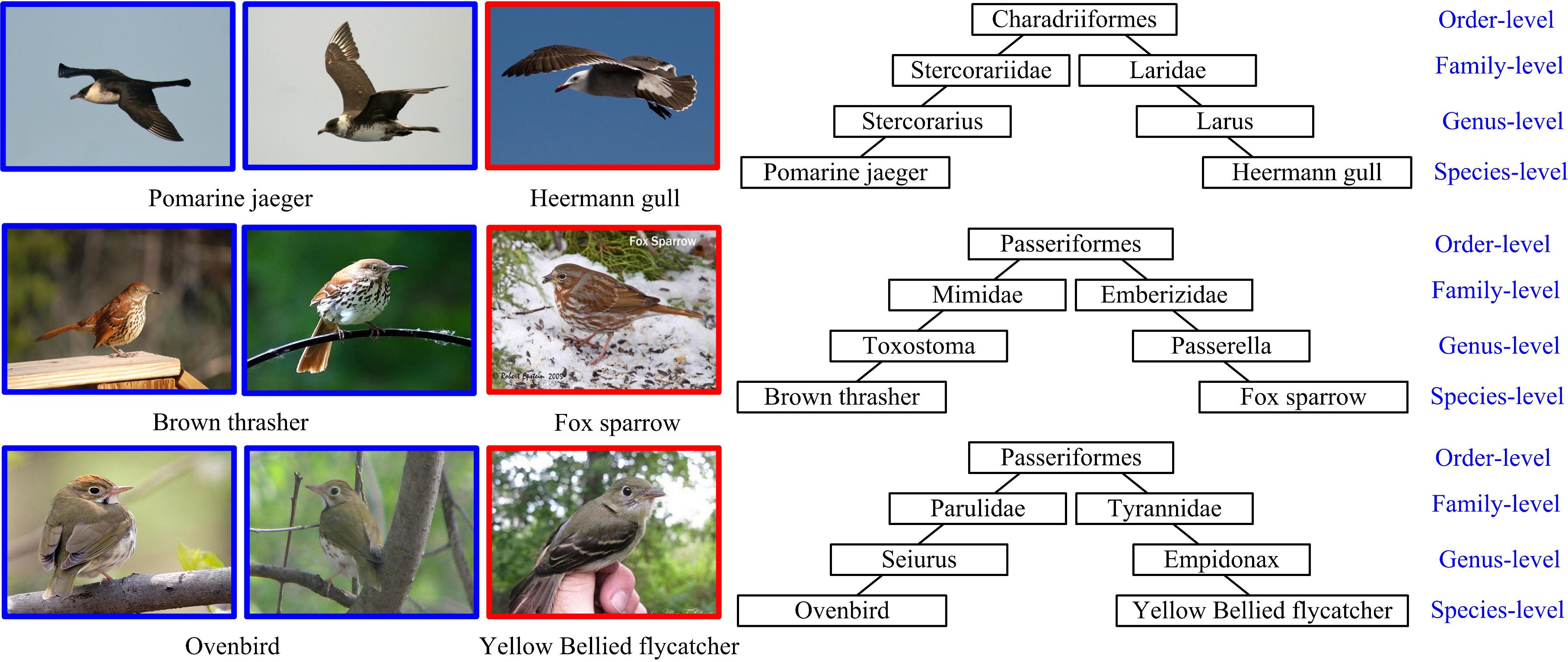}
\end{center}
\vspace{-0.05in}
\caption{Samples of misclassification only with species/genus-level features. In this figure, images in the blue boxes are misclassified images only with species/genus-level features and the class names under images are their true labels, while the predicted labels of these images (in blue boxes) are given by a sample (in red boxes) in the corresponding rows. These misclassified images are correctly classified when all species/genus/family-level features are used.}
\label{mis2}
\end{figure*}

Note that the labels of training data do not include unseen classes
and thus domain shift will occur when we extract features for test
images using the deep neural networks trained by these training
data \cite{Kodirov15}. To avoid domain shift, we add a domain adaptation structure \cite{Ganin15}, which includes a gradient reversal layer and a domain classifier, after the fifth group of convolutional layers in VGG-16Net (as shown in Fig. \ref{flowchart}). The domain adaption structure views training data and test data as two domains and aims to train a domain classifier that cannot distinguish its domain of a given data. In this way, the difference of features among data from
two domains can be eliminated. In our model, we aim to achieve an
adversarial process, i.e. to learn features that can confuse the
domain classifier and classify fine-grained classes. Therefore, we
aim to minimize the loss of hierarchical classification subnetworks
and maximize the loss of the domain classifier. The gradient
reversal layer (Grl layer in Fig. \ref{dc}) proposed by
\cite{Ganin15} is used to achieve the goal. In the following, we denote the domain classifier as $G_d$, which is also presented in Fig. \ref{dc}. By merging the domain adaptation structure, hierarchical classification subnetworks and VGG-16Net together, we define the total loss for image $x$ as:
\begin{equation}
\begin{split}
\mathcal {L}(\theta_F,\theta_f,\theta_g,\theta_s,\theta_d)&=\mathcal{L}_h(\theta_F,\theta_f,\theta_g,\theta_s)-\\
&\mu_d\mathcal{L}_d(y_d,G_d(G_s(\theta_s,G(x;\theta_F));\theta_d))
\end{split}
\end{equation}
where $\mathcal {L}_d$, $y_d$, $\mu_d$ and $\theta_d$ denote the loss of domain classifier, the domain label of image $x$, the weight of loss of domain classifier, and the parameters of domain classifier, respectively.

To end this subsection, we qualitatively demonstrate the important role of the hierarchical semantic structure of fine-grained classes in extracting discriminative features for zero-shot fine-grained classification. Fig.~\ref{mis1} provides some samples of misclassified images when only species-level features are used, and Fig.~\ref{mis2} provides some samples of misclassified images when only species/genus-level features are used. It can be seen that the true labels and predicted labels of these misclassified images (in blue boxes) have hierarchical semantical relations, and these misclassified images can be correctly classified when higher-level features are used. That is, the hierarchical semantic structure of fine-grained classes can be used to capture more discriminative features.

\subsection{Label Inference}
\label{sect:zsl_method}

In this subsection, with the discriminative features obtained from
Section \ref{sect:fea_learn}, we provide a label propagation
approach for zero-shot fine-grained image classification.

Let $S=\{s_1,...,s_p\}$ denote the set of seen classes and
$U=\{u_1,...,u_q\}$ denote the set of unseen classes, where $p$ and
$q$ are the total numbers of seen classes and unseen classes,
respectively. These two sets of classes are disjoint, i.e. $S\cap
U=\phi$. We are given a set of labeled training images
$D_s=\{(x_i,y_i ):i=1,...,N_s\}$, where $x_i$ is the feature vector
of the $i$-th image in the training set, $y_i\in S$ is the
corresponding label, and $N_s$ denotes the total number of labeled
images. Let $D_u=\{(x_j,y_j ):j=1,...,N_u\}$ denote a set of
unlabeled test images, where $x_j$ is the feature vector of the
$j$-th image in the test set, $y_j\in U$ is the corresponding
unknown label, and $N_u$ denotes the total number of unlabeled
images. The main goal of zero-shot learning is to predict $y_j$ by
learning a classifier $f:X_u\rightarrow U$, where
$X_u=\{x_j:j=1,...,N_u\}$.

For zero-shot fine-grained image classification, we first need
estimate the semantic relationships between seen and unseen classes,
which will be used for predicting the labels of images in unseen
classes. In this paper, we collect the attributes of each fine-grained class to form its semantic vector and further construct a semantic-directed graph $\mathcal {G}=\{V,E\}$ over all the classes (including seen/unseen), where $V$ denotes the set of nodes (i.e. fine-grained classes) in the graph and $E$ denotes the set of directed edges between classes. The detailed steps of constructing the graph $\mathcal {G}$ is given as follows:
\begin{itemize}
\item
We first construct the edges between seen classes. Specifically,
for each seen class, we perform the $k$-nearest-neighbors ($k$-NN) method on semantic vectors to find its $k_1$ nearest neighbors in seen classes, and then construct a directed edge from this seen class to each of its neighbors. The weight of this edge is the negative exponent of the Euclidean distance between them.
\item
We further take the same strategy to construct edges between seen classes and unseen classes. Specifically, for each seen class, we perform the $k$-NN method on semantic vectors to find its $k_2$ nearest neighbors in unseen classes, and then construct a directed edge from this seen class to each of its neighboring unseen classes. The weight of this edge is the negative exponent of the Euclidean distance between them.
\item
Finally, for each unseen classes, it has only one edge pointing to itself with a weight of 1.
\end{itemize}

\begin{algorithm}[t]
   \caption{The Proposed Framework}
   \label{alg1}
\begin{algorithmic}
   \STATE {\bfseries Input:} the set of labeled training images $D_s$ \\ \qquad ~~ the set of test images in unseen classes $X_u$\\
   \STATE {\bfseries Feature Learning:}
   \STATE {\bfseries 1)} Train the proposed neural network using hierarchical semantic structure among fine-grained classes;
   \STATE {\bfseries 2)} Run forward computation of the proposed neural network for each test image and extract deep features from hierarchical classification subnetworks;
   \STATE {\bfseries 3)} Concatenate the features from hierarchical classification subnetworks to obtain deep features $F$;
   \STATE {\bfseries Label Inference:}
   \STATE {\bfseries 4)} Compute the initial probabilities of test images belonging to unseen classes $Y$ with the LIBLINEAR toolbox \cite{RFan} and deep features ${F}$;
   \STATE {\bfseries 5)} Construct the semantic-directed graph based on semantic vectors;
   \STATE {\bfseries 6)} Compute the normalized transition matrix $P$ according to Equations (3-5);
   \STATE {\bfseries 7)} Find the solution $\tilde{Y}^*$ of label propagation problem formulated in Equation (6) according to Equations (7-8);
   \STATE {\bfseries 8)} Label each test image $x_i$ with class $\arg\max_j \tilde{Y}^*_{ij}$.
   \STATE {\bfseries Output:} Labels of test images in unseen classes.
\end{algorithmic}
\end{algorithm}

By collecting the above edge weights up, we can denote the weight
matrix ${W}$of the semantic-directed graph
$\mathcal {G}$ as:
\begin{equation}
{W}=\begin{bmatrix}R_1&R_2
\\0&I\end{bmatrix}
\end{equation}
where $R_1\in \mathbb {R}^{p\times p}$ collects the edge weights among seen classes, $R_2\in \mathbb{R}^{p \times q}$ collects the edge weights between seen classes and unseen classes, and $I\in \mathbb {R}^{q\times q }$ denotes an identity matrix. A Markov chain process can be further defined over the graph $\mathcal {G}$ by constructing the transition matrix:
\begin{equation}
{T}={D}^{-1}{W}
\end{equation}
where ${D}$ is a diagonal matrix with its $i$-th diagonal element being equal to the sum of the $i$-th row of ${W}$.

To guarantee that the Markov chain process has a unique stationary
solution \cite{Zhou}, we normalize ${T}$ as:
\begin{equation}
P= \frac{\eta}{p+q-1}(1_{p+q}-I_{p+q})+(1-\eta){T}
\end{equation}
where $\eta$ is a normalization parameter (empirically set as $\eta=0.001$), and $1_{p+q}$ and $I_{p+q}$ are the one matrix and
identity matrix of the size $(p+q)\times (p+q)$, respectively.

Based on the normalized transition matrix $P= [p_{uv}] \in
\mathbb{R}^{(p+q)\times(p+q)}$, we formulate zero-shot fine-grained
classification as the following label propagation problem:
\begin{equation}
\min\limits_{\tilde{Y}_{i.}}\frac{1}{2}\sum_{u,v}
\pi(u)p_{uv}(\frac{\tilde{Y}_{iu}}{\sqrt{\pi(u)}}- \frac{\tilde{Y}_{iv}}{\sqrt{\pi(v)}})^2+\lambda\|\tilde{Y}_{i.}-Y_{i.}\|_2^2
\end{equation}
where $\tilde{Y}_{i.}$ (the $i$-th row of $\tilde{Y} \in
\mathbb{R}^{N_u\times (p+q)}$) and $Y_{i.}$ (the $i$-th row of $Y\in
\mathbb{R}^{N_u\times (p+q)}$) collect the optimal and initial
probabilities of the $i$-th test image belonging to each class,
respectively. That is, $Y$ is an initialization of $\tilde{Y}$ and
$\tilde{Y}$ is the final solution of the problem formulated in
Equation (6). Moreover, $\pi(u)$ is the sum of the $u$-th row of $P$
(i.e. $\sum_{v}p_{uv}$), and $\lambda$ is a regularization
parameter.

The first term of the above objective function sums the weighted
variation of $\tilde{Y}_{i.}$ on each edge of the directed graph
$\mathcal {G}$, which aims to ensure that $\tilde{Y}_{i.}$ does not
change too much between semantically similar classes for the $i$-th
test image. The second term denotes an $L_2$-norm fitting
constraint, which means that $\tilde{Y}_{i.}$ should not change too
much from $Y_{i.}$.

To solve the above label propagation problem, we adopt the technique
introduced in \cite{Zhou} and define the operator $\Theta$:
\begin{equation}
\Theta=(\Pi^{1/2}P\Pi^{-1/2}+\Pi^{-1/2}P\Pi^{1/2})/2
\end{equation}
where $\Pi$ is a $(p+q)\times (p+q)$ diagonal matrix with its $u$-th
diagonal element being equal to $\pi(u)$. According to \cite{Zhou}, the optimal solution $\tilde{Y}^*$ of the problem in Equation (6) is:
\begin{equation}
\tilde{Y}^*=Y(I-\alpha \Theta)^{-1}
\end{equation}
where $I \in \mathbb{R}^{(p+q)\times(p+q)}$ denotes an identity
matrix and $\alpha=1/(1+\lambda)\in (0,1)$.

\begin{figure}[t]
\vspace{0.02in}
\begin{center}
\includegraphics[width=0.93\columnwidth]{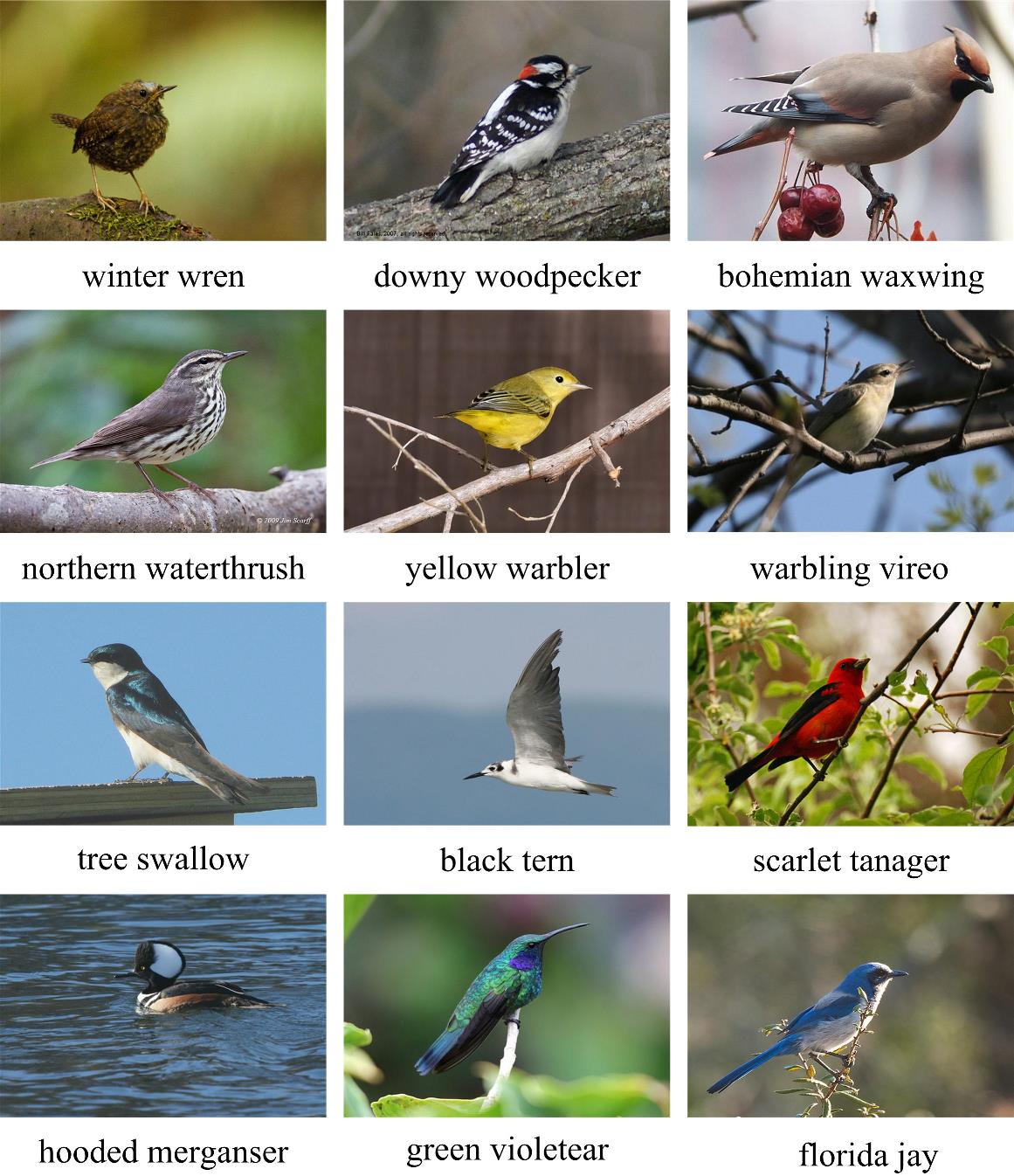}
\end{center}
\caption{Examples of images in the Caltech UCSD Birds-200-2011
Dataset. Corresponding categories are given below images.}
\label{CUB}
\end{figure}

\begin{figure}[t]
\vspace{0.02in}
\begin{center}
\includegraphics[width=0.93\columnwidth]{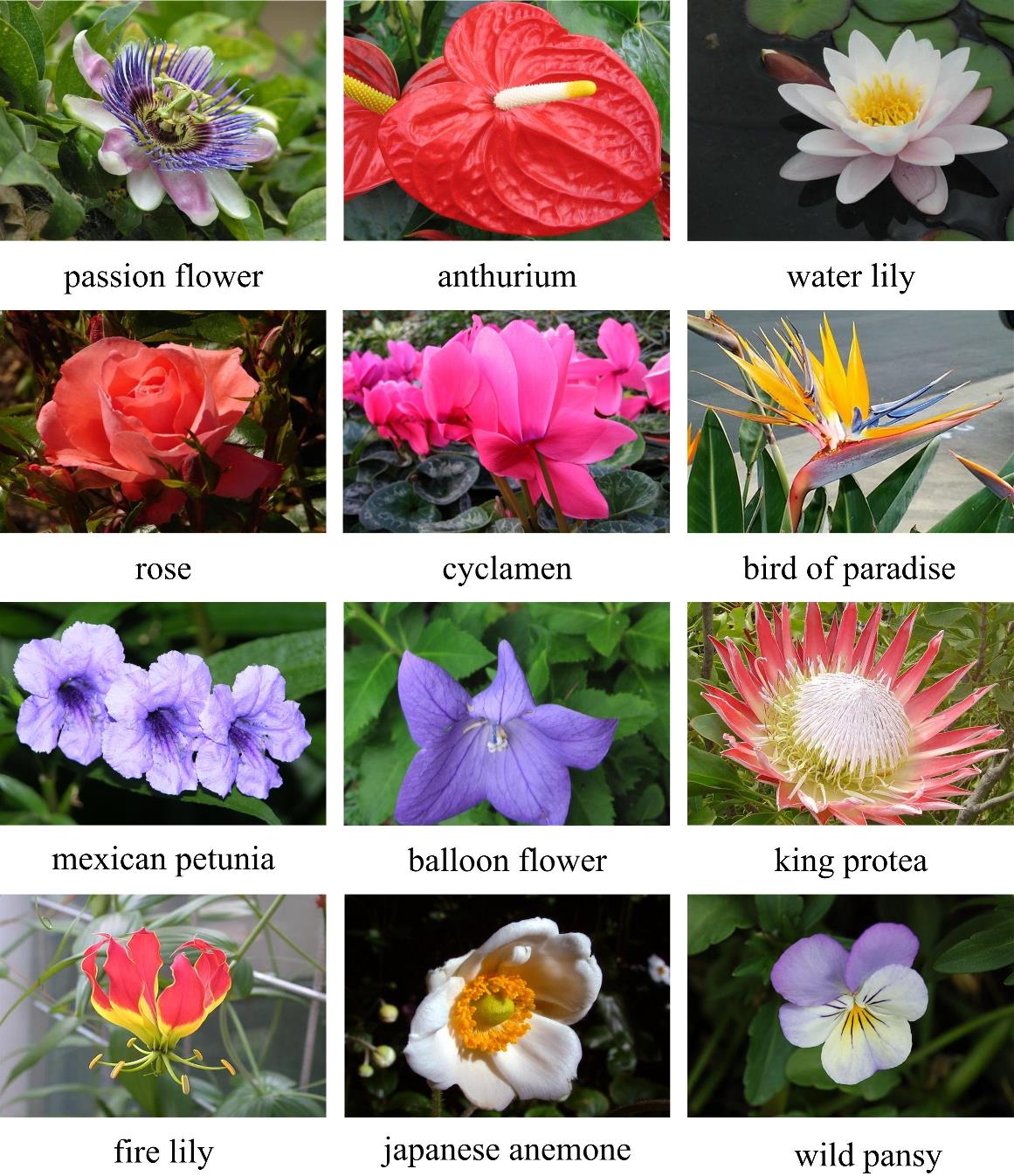}
\end{center}
\caption{Examples of images in the Oxford Flower-102 Dataset.
Corresponding categories are given below images.} \label{OF}
\end{figure}

To obtain the above solution, we need to provide $Y$ in advance.
Note that each row of $Y$ consists of two parts: the probabilities
of a test image belonging to seen classes, and the probabilities of
a test image belonging to unseen classes. Given no labeled data in
unseen classes, we directly set the probabilities belonging to
unseen classes as 0. To compute the initial probabilities belonging
to seen classes, we use LIBLINEAR toolbox \cite{RFan} to train an
$L_2$-regularized logistic regression classifier. In general, we
empirically set the parameter $c$ in $L_2$-regularized logistic
regression as 0.01.

To sum up, by combining the coarse-to-fine feature learning and label
propagation approaches together, the complete algorithm for zero-shot
fine-grained classification is outlined as Algorithm~\ref{alg1}. It
should be noted that the proposed approach can be extended to weak
supervision setting by replacing class attributes with semantic
vectors extracted by word vector extraction methods (e.g. word2vec
\cite{word2vec}).

\begin{table*}[t]
\centering
\caption{Comparison of the concatenation of different-level features on CUB-200-2011. Here, `FTVGG-16+HCS' and `FTVGG-16+HCS+DA' denote features obtained from finetuned VGG-16Net with hierarchical classification subnetworks, and finetuned VGG-16Net with hierarchical classification subnetworks and domain adaptation structure, respectively. }
 \label{CUB_fea} \tabcolsep0.8cm
 \begin{tabular}{c|l|c|c}
\hline
\multirow{2}{*}{Features}&\multirow{2}{*}{Semantic Level}&\multicolumn{2}{c}{Accuracy ($\%$)}\\
\cline{3-4}&&Class Attributes &Semantic Vectors\\\hline
 \multirow{5}{*}{FTVGG-16+HCS}& Species-level & 44.9 &  28.9\\
 & Genus-level & 36.3 &  22.3\\
 & Family-level & 32.3 &  15.3\\
 & Species/Genus-level & 45.7 &  30.4\\
 & Species/Genus/Family-level & \textbf{46.2} & \textbf{32.2}\\
 \hline\vspace{0.01in}
 \multirow{5}{*}{FTVGG-16+HCS+DA}& Species-level & 46.8 & 29.8\\
 & Genus-level & 37.1 &  24.3\\
 & Family-level & 33.2 &  18.3\\
 & Species/Genus-level & 48.3 &  33.2\\
 & Species/Genus/Family-level & \textbf{49.5} &  \textbf{34.5}\\\hline
\end{tabular}
\end{table*}

\begin{figure*}[t]
\vspace{0.05in} \centering
\includegraphics[width=0.93\textwidth]{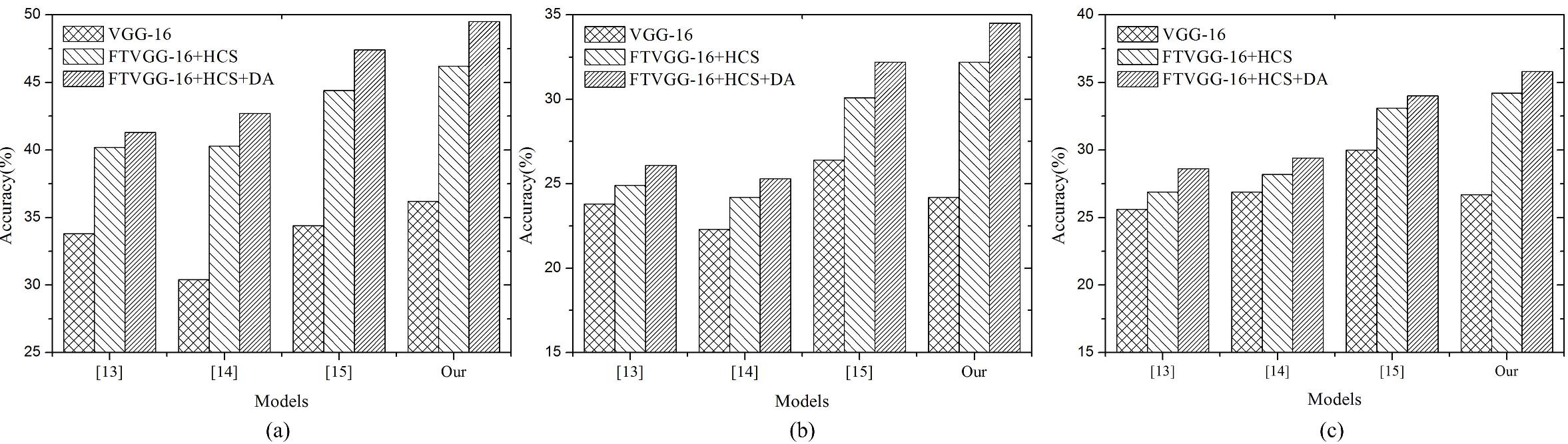}
\vspace{-0.05in} \caption{The results of different zero-shot learning
models using the proposed feature learning approach. (a) CUB-200-2011 with class attributes. (b) CUB-200-2011 with semantic vectors. (c) Flowers-102 with semantic vectors. In this figure, `VGG-16', `FTVGG-16+HCS' and `FTVGG-16+HCS+DA' denote features obtained from VGG-16Net pretrained by ImageNet, finetuned VGG-16Net with hierarchical classification subnetworks and finetuned VGG-16Net with hierarchical classification subnetworks and domain adaptation structure, respectively. } \label{CUB_res_fe2}
\end{figure*}

\section{Experimental Results and Discussions}

\subsection{Experimental Setup}

We conduct experiments on two benchmark fine-grained datasets, i.e. Caltech UCSD Birds-200-2011 \cite{CUB-200-2011} and Oxford Flower-102 \cite{Flower}. Our experimental setup is give as follows.

\subsubsection{Caltech UCSD Birds-200-2011 Dataset}

The Caltech UCSD Birds -200-2011 dataset (CUB-200-2011) contains
11,788 images of 200 North-American bird species
\cite{CUB-200-2011}. Each species is associated with a Wikipedia
article and organized by scientific classification (family, genus,
species). Each class is also annotated with 312 visual attributes.
In the zero-shot setting, we follow \cite{Zhang16} to use 150 bird
spices as seen classes for training and the left 50 spices as unseen
classes for testing. The results are the average of four fold cross
validation. For parameter validation, we also use a zero-shot
setting within the 150 classes of the training set, i.e. we
use 100 classes for training and the rest for validation. The
hierarchical labels of fine-grained classes are collected from
Wikipedia. For each fine-grained class, we use 312-d class
attributes as semantic description, or 300-d semantic vectors extracted by the word2vec model \cite{word2vec} (trained by GoogLeNews).
Examples of images in CUB-200-2011 are shown in Fig. \ref{CUB}.

\subsubsection{Oxford Flower-102 Dataset}

The Oxford Flower-102 (Flowers-102) dataset contains 8,189 images of
102 different categories. There is no human annotated attribute for
each category. Therefore, we only choose 80 of 102 categories, to ensure that each selected category is associated with a Wikipedia article and organized by scientific classification (family, genus, species). In the zero-shot setting, similar to the setting of CUB-200-2011, we use 60 flower spices as seen classes for training and the left 20 spices as unseen classes for testing. The results are the average of four fold cross validation. For parameter validation, we use similar strategy as CUB-200-2011, i.e. we use 40 classes for training and the rest for validation. The hierarchical labels of fine-grained classes are collected from Wikipedia. For each fine-grained class,
only 300-d semantic vectors extracted by the word2vec model
\cite{word2vec} (trained by GoogLeNews) are used as semantic
description. Examples of Flower-102 are shown in Fig.~\ref{OF}.

\begin{table*}[t]
\centering
 \caption{Comparison of different zero-shot fine-grained classification approaches with class attributes.}
\label{CUB_res_1} \tabcolsep0.6cm
\begin{tabular}{c|l|l|c}
\hline Datasets &Approaches & Features & Accuracy ($\%$)\\
\hline \multirow{9}{*}{CUB-200-2011}
 &~\cite{Paredes15}~& VGG-16Net & 33.8\\
 &~\cite{Zhang15}~& VGG-16Net & 30.4\\
 &~\cite{Fu15}~& VGG-16Net & 34.4\\
 &~\cite{Zhang16}~& VGG-16Net & 42.1\\
 &~\cite{Bucher16}~& VGG-16Net & 43.3 \\
 &~\cite{Akata16}~& VGG-16Net & 43.3\\
 &~ZC~& VGG-16Net & 36.2\\
 &~Ours-1~& Finetuned VGG-16Net+HCS& 46.2 \\
 &~Ours-2~& Finetuned VGG-16Net+HCS+DA& \textbf{49.5} \\
\hline
Flowers-102 &~--~& -- & -- \\
\hline
\end{tabular}
\end{table*}

\begin{table*}[t]
\centering
 \caption{Comparison of different zero-shot fine-grained classification approaches with semantic vectors.}
\label{CUB_res_2} \tabcolsep0.6cm
\begin{tabular}{c|l|l|c}
\hline Datasets &Approaches & Features & Accuracy ($\%$)\\
\hline \multirow{7}{*}{CUB-200-2011} &~\cite{Paredes15}~& VGG-16Net & 23.8 \\
 &~\cite{Zhang15}~& VGG-16Net  & 22.3\\
 &~\cite{Fu15}~& VGG-16Net & 26.4 \\
 &~\cite{Qiao16}~& VGG-16Net & 29.0 \\
 &~ZC~&VGG-16Net& 24.2\\
 &~Ours-1~& Finetuned VGG-16Net+HCS& 32.2 \\
 &~Ours-2~& Finetuned VGG-16Net+HCS+DA& \textbf{34.5} \\
 \hline \multirow{6}{*}{Flowers-102}&~\cite{Paredes15}~& VGG-16Net & 25.6 \\
 &~\cite{Zhang15}~& VGG-16Net & 27.3\\
 &~\cite{Fu15}~& VGG-16Net & 30.8 \\
 &~ZC~&VGG-16Net& 26.7\\
 &~Ours-1~& Finetuned VGG-16Net+HCS& 34.2 \\
 &~Ours-2~& Finetuned VGG-16Net+HCS+DA& \textbf{35.8} \\
 \hline
\end{tabular}
\end{table*}

\subsection{Implementation Details}

In the feature learning phase, the VGG-16Net's layers are pre-trained
on ILSVRC 2012 1K classification \cite{Russakovsky}, and then
finetuned with the training data. Meanwhile, the other layers are trained from scratch. All input images are resized to 224$\times$224 pixels. Stochastic gradient descent (SGD)\cite{sgd} is used to optimize our model with a basic learning rate of 0.01, a momentum of 0.9, a
weight decay of 0.005 and a minibatch size of 20. For layers trained
from scratch, their learning rate is 10 times of basic learning
rate. Our model is implemented based on the popular Caffe \cite{jia2014caffe}.

Note that different-level features are extracted from the last
but one fully-connected layers before the softmax layers. Hence, we finally obtain three kinds of features which are used to classify images at different levels. To find a good way to combine these features, we conduct experiments with the proposed model using the concatenation of different-level features. The results are shown in Table~\ref{CUB_fea}. It can be seen that high-level features perform better than features extracted from the shallow layers. Furthermore, we also find that the combination of three-level features performs the best. In the following, we use the concatenation of three-level features as the final deep visual features in our model.

\subsection{Effectiveness of Deep Feature Learning}

To test the effectiveness of the proposed feature learning approach,
we apply the features extracted by the the proposed feature learning
approach to different zero-shot learning models (e.g. \cite{Fu15,Zhang15,Paredes15}) under the same setting. The results are given in Fig.~\ref{CUB_res_fe2}. It can be seen that the proposed feature learning approach works well in all the zero-shot learning models. This observation can be explained as follows. Compared with the traditional VGG-16Net pretrained by ImageNet, the proposed feature learning approach takes hierarchical semantic structure of classes and domain adaptation structure into account, and thus succeeds in generating more discriminative features for zero-shot fine-grained classification.

\subsection{Comparison to the State-of-the-Arts}

\subsubsection{Test with Class Attributes}

We provide the comparison of the proposed approach to the
state-of-the-art zero-shot fine-grained classification approaches
\cite{Akata16,Bucher16,Zhang15,Zhang16,Fu15,Paredes15} using class
attributes, which is shown in Table~\ref{CUB_res_1}. Since Flowers-102 provides no class attributes, we do not present the results on this dataset. In this table, `ZC' denotes the zero-shot learning approach based on label propagation, `VGG-16Net' denotes the features obtained from VGG-16Net \cite{Simonyan15} (pretrained with ImageNet \cite{Russakovsky}), `HCS' denotes the hierarchical classification subnetworks proposed in Section~\ref{sect:fea_learn}, and `DA' denotes the domain adaptation structure proposed in Section~\ref{sect:fea_learn}. It can be seen that the proposed approach significantly outperforms the state-of-the-art zero-shot learning approaches. That is, both feature learning and label inference used in the proposed approach are crucial for zero-shot fine-grained classification. Moreover, the comparison between `ZC' vs. `Ours-2' demonstrates that the proposed feature learning approach is extremely effective in the task of zero-shot fine-grained classification. Additionally, the comparison between `Ours-1' vs. `Ours-2' demonstrates that the domain adaptation structure is important for feature learning in the task of zero-shot fine-grained classification. It should be noted that \cite{Akata16} has achieved an accuracy of 56.5 $\%$ using a multi-cue framework, where locations of parts are used as very strong supervision in both training and test process. The accuracy of 49.5 $\%$ released in Table \ref{CUB_res_1} is its classification result when only annotations of the whole images are used (without locations of parts). The superior performance of the proposed approach compared with \cite{Akata16} further verifies the effectiveness of the proposed approach in zero-shot fine-grained classification.

\subsubsection{Test with Semantic Vectors}

We also evaluate the proposed approach in the weakly supervised
setting, where only fine-grained labels of training images are given
and the semantics among fine-grained are learned from text
descriptions. Table \ref{CUB_res_2} provides the classification results on both CUB-200-2011 and Flowers-102 in the weaker supervised
setting. From this table, we can still observe that the proposed approach outperforms the state-of-the-art approaches, which further verifies the effectiveness of the proposed approach.

\section{Conclusion}

In this paper, we propose a two-phase framework for zero-shot
fine-grained classification approach, which can recognize images
from unseen fine-grained classes. In our approach, a feature
learning strategy based on the hierarchical semantic structure of
fine-grained classes and domain adaptation structure is developed to generate robust and discriminative features, and then a label propagation method based on semantic directed graph is proposed for label inference. Experimental results on the benchmark fine-grained
classification datasets demonstrate that the proposed approach
outperforms the state-of-the-art zero-shot learning algorithms. Our
approach can also be extended to the weakly supervised setting (i.e. only fine-grained labels of training images are given) and has achieved better results than the state-of-the-arts. In the future work, we
will make further improvements on developing more powerful word
vector extractors to explore better semantic relationships among
fine-grained classes and optimize the feature extractors with word
vector extractors simultaneously.

\section*{Acknowledgements}

This work was partially supported by National Natural Science Foundation of China (61573363 and 61573026), 973 Program of China (2014CB340403 and 2015CB352502), the Fundamental Research Funds for the Central Universities and the Research Funds of Renmin University of China (15XNLQ01), and European Research Council FP7 Project SUNNY (313243).

\IEEEtriggeratref{21}

\end{document}